\documentclass[letterpaper, 10 pt, conference]{ieeeconf}  

\IEEEoverridecommandlockouts                              

\usepackage{graphics} 
\usepackage{graphicx} 
\usepackage{amsmath} 
\usepackage{amssymb}  
\usepackage{mathtools}

\usepackage{subcaption}
\usepackage{booktabs}

\usepackage{array}

\newcolumntype{P}[1]{>{\raggedright\arraybackslash}p{#1}}

\usepackage{tikz}
\usetikzlibrary{bayesnet}

\usetikzlibrary{arrows.meta}

\definecolor{color1bg}{HTML}{458588}
\tikzset{%
  >={Latex[width=2mm,length=2mm]},
            base/.style = {rectangle, rounded corners, draw=black,
                           minimum width=1cm, minimum height=1cm,
                           text centered, font=\small},
  activityStarts/.style = {base, fill=blue!30},
       startstop/.style = {base, fill=red!30},
    activityRuns/.style = {base, fill=green!30},
         inout/.style = {base, minimum width=6cm, minimum height=0cm,
                           font=\small},
         process/.style = {base, minimum width=1.5cm, 
                           font=\small},
         data/.style = {base,minimum width=2.8cm, 
                      minimum height=2.8cm, 
                           font=\small},
         sec/.style = {base, minimum width=1.5cm, 
                           font=\small, draw=none,
                           fill=none },
         processX/.style = {base, minimum width=1.8cm, 
                           inner sep=1pt,
                           minimum height=.5cm,
                           font=\footnotesize},
        gnn_var/.style = {circle, draw=black,minimum height=.2cm,
           text centered, font=\small},
        gnn_con/.style = {rectangle, draw=black,minimum height=.3cm,
minimum width=.3cm,
                       text centered, font=\small, fill=color1bg},
         processR/.style = {processX, fill=red!30},
         processG/.style = {processX, fill=green!30},
         number/.style = { minimum width=1cm, minimum height=0cm,
                           text centered, font=\small},
}

\usepackage[ruled,vlined,linesnumbered]{algorithm2e}
\SetInd{0.25em}{0.5em}
\SetAlFnt{\footnotesize\sf}
\SetKwRepeat{Do}{do}{while}
\SetKw{KwStep}{step}
\SetKwInput{KwData}{Input}
\SetKwInput{KwResult}{Result}
\SetKwBlock{RepeatForever}{repeat}{}
\SetKw{Continue}{continue}
\SetKwComment{Comment}{$\triangleright$\ }{}
\SetCommentSty{itshape}

\newcommand{\R}{\mathbb{R}}

\usepackage{etoolbox}
\makeatletter
\patchcmd{\@makecaption}
  {\scshape}
  {}
  {}
  {}
\makeatother

\graphicspath{{./images/manip_seq/crop/}{./images/vid_new/}}

\title{\LARGE \bf

  Learning Feasibility of Factored Nonlinear Programs in Robotic Manipulation Planning

}

\author{
  Joaquim Ortiz-Haro$^{1}$, Jung-Su Ha$^{1}$,
  Danny Driess$^{1}$,
  Erez Karpas$^{2}$,
  Marc Toussaint$^{1}$
\thanks{$^{1}$ TU Berlin, Germany. \quad $^{2}$  Technion, Israel. }
\thanks{This research has been supported by the German-Israeli Foundation for Scientific Research (GIF) grant I-1491-407.6/2019. Joaquim Ortiz-Haro and Danny Driess thank the International Max-Planck Research School for Intelligent Systems (IMPRS-IS) for the support.}
}

\begin{document}

\maketitle
\thispagestyle{empty}
\pagestyle{empty}

\begin{abstract}


A factored Nonlinear Program (Factored-NLP) explicitly models the dependencies between a set of continuous variables and nonlinear constraints, providing an expressive formulation for relevant robotics problems such as manipulation planning or simultaneous localization and mapping.
When the problem is over-constrained or infeasible, a fundamental issue is to detect a minimal subset of variables and constraints that are infeasible. 
Previous approaches require solving several nonlinear programs, incrementally adding and removing constraints, and are thus computationally expensive.
In this paper, we propose a graph neural architecture that predicts which variables and constraints are jointly infeasible.
The model is trained with a dataset of labeled subgraphs of Factored-NLPs, and importantly, can make useful predictions on larger factored nonlinear programs than the ones seen during training.
We evaluate our approach in robotic manipulation planning, where our model is able to generalize to longer manipulation sequences involving more objects and robots, and different geometric environments. 
The experiments show that the learned model accelerates general algorithms for conflict extraction (by a factor of 50) and heuristic algorithms that exploit expert knowledge (by a factor of 4).

\end{abstract}

\section{INTRODUCTION}


Computing values for a set of variables that fulfil all the constraints is a key problem in several applications, such as robotics, planning, and scheduling.
In discrete domains, these problems are generally known as Constrained Satisfaction Problems (CSP), which also include classical combinatorial optimization like $k$-coloring, maximum cut or Boolean satisfaction (SAT).
In continuous domains, the dependencies between a set of continuous variables and nonlinear constraints can be modelled with a factored nonlinear program (without cost term or  with a small regularization), which have applications in manipulation planning \cite{toussaint2018differentiable} or simultaneous localization and mapping \cite{dellaert2017factor}.




When a problem is over-constrained or infeasible, a fundamental challenge is to extract a minimal conflict: a minimal subset of variables and constraints that are jointly infeasible. These conflicts usually provide an explanation of the failure that can be incorporated back into iterative solvers, for example in the conflict-driven clause learning algorithm for SAT problems \cite{bayardo1997using, marques1999grasp}, or conflict based solvers for Task and Motion Planning in robotics (TAMP) 
\cite{dantam2018incremental, srivastava2014combined, Ortiz2022Conflict-Directed, ortiz2022conflict, garrett2021integrated}.




\begin{figure}
  \centering

\begin{subfigure}[c]{0.49\columnwidth}

\begin{tikzpicture}[node distance=1.2cm,
    every node/.style={fill=white, font=\small}, align=center]
  \node (var1)     [gnn_var]          {$$};
  \node (var2)     [right of=var1, xshift=3em,  gnn_var]          {$$};

  \node (var3)     [below of=var1, gnn_var]          {$$};
  \node (var4)     [below of=var2, gnn_var]          {$$};
  \node (var5)     [below of=var3, gnn_var]          {$$};
  \node (var6)     [below of=var4, gnn_var]          {$$};

  \node (con12)     [right of = var1 , xshift=-.5em,gnn_con ]          {$$};
  \node (con13)     [below of = var1 , yshift=2em, xshift=-2em, gnn_con]          {$$};
  \node (con35)     [below of = var3 , yshift=2em, xshift=-2em, gnn_con]          {$$};
  \node (con234)     [below of = con12 ,yshift=1.5em, gnn_con]          {$$};

  \node (con546)     [below of = con234 , yshift=1em, gnn_con]          {$$};

  \node (con56)     [right of = var5 , gnn_con]          {$$};

  \draw[-]             (con12) -- (var1);
  \draw[-]             (con12) -- (var2);

  \draw[-]             (con234) -- (var3);
  \draw[-]             (con234) -- (var4);
  \draw[-]             (con234) -- (var2);

  \draw[-]             (con546) -- (var5);
  \draw[-]             (con546) -- (var4);
  \draw[-]             (con546) -- (var6);

  \draw[-]             (con13) -- (var3);
  \draw[-]             (con13) -- (var1);

  \draw[-]             (con35) -- (var3);
  \draw[-]             (con35) -- (var5);

  \draw[-]             (con56) -- (var5);
  \draw[-]             (con56) -- (var6);

\end{tikzpicture}
\subcaption{Input Factored-NLP}
\end{subfigure}
\begin{subfigure}[c]{0.49\columnwidth}
\begin{tikzpicture}[node distance=1.2cm,
    every node/.style={fill=white, font=\small}, align=center]
  \node (var1)     [gnn_var]          {$$};
  \node (var2)     [right of=var1, xshift=3em,  gnn_var]          {$$};

  \node (var3)     [below of=var1, gnn_var]          {$$};
  \node (var4)     [below of=var2, gnn_var]          {$$};
  \node (var5)     [below of=var3, gnn_var]          {$$};
  \node (var6)     [below of=var4, gnn_var]          {$$};

  \node (con12)     [right of = var1 , xshift=-.5em,gnn_con ]          {$$};
  \node (con13)     [below of = var1 , yshift=2em, xshift=-2em, gnn_con]          {$$};
  \node (con35)     [below of = var3 , yshift=2em, xshift=-2em, gnn_con]          {$$};
  \node (con234)     [below of = con12 ,yshift=1.5em, gnn_con]          {$$};

  \node (con546)     [below of = con234 , yshift=1em, gnn_con]          {$$};

  \node (con56)     [right of = var5 , gnn_con]          {$$};

  \draw[latex-latex]             (con12) -- (var1);
  \draw[latex-latex]             (con12) -- (var2);

  \draw[latex-latex]             (con234) -- (var3);
  \draw[latex-latex]             (con234) -- (var4);
  \draw[latex-latex]             (con234) -- (var2);

  \draw[latex-latex]             (con546) -- (var5);
  \draw[latex-latex]             (con546) -- (var4);
  \draw[latex-latex]             (con546) -- (var6);

  \draw[latex-latex] (con13) -- (var3);
  \draw[latex-latex]             (con13) -- (var1);

  \draw[latex-latex]             (con35) -- (var3);
  \draw[latex-latex]             (con35) -- (var5);

  \draw[latex-latex]             (con56) -- (var5);
  \draw[latex-latex]             (con56) -- (var6);

\end{tikzpicture}

\caption{Neural message passing}
\end{subfigure}

\vspace{.5cm}











\begin{subfigure}[c]{0.49\columnwidth}
\begin{tikzpicture}[node distance=1.2cm,
    every node/.style={fill=white, font=\small}, align=center]
  \node (var1)     [gnn_var, fill=darkgray]          {$$};
  \node (var2)     [right of=var1, xshift=3em,  gnn_var, fill=black]          {$$};

  \node (var3)     [below of=var1, gnn_var, fill=lightgray]          {$$};
  \node (var4)     [below of=var2, gnn_var, fill=lightgray]          {$$};
  \node (var5)     [below of=var3, gnn_var, fill=black]          {$$};
  \node (var6)     [below of=var4, gnn_var, fill=gray]          {$$};

  \node (con12)     [right of = var1 , xshift=-.5em,gnn_con]          {$$};
  \node (con13)     [below of = var1 , yshift=2em, xshift=-2em, gnn_con]          {$$};
  \node (con35)     [below of = var3 , yshift=2em, xshift=-2em, gnn_con]          {$$};
  \node (con234)     [below of = con12 ,yshift=1.5em, gnn_con]          {$$};

  \node (con546)     [below of = con234 , yshift=1em, gnn_con]          {$$};

  \node (con56)     [right of = var5 , gnn_con]          {$$};

  \draw[-]             (con12) -- (var1);
  \draw[-]             (con12) -- (var2);

  \draw[-]             (con234) -- (var3);
  \draw[-]             (con234) -- (var4);
  \draw[-]             (con234) -- (var2);

  \draw[-]             (con546) -- (var5);
  \draw[-]             (con546) -- (var4);
  \draw[-]             (con546) -- (var6);

  \draw[-]             (con13) -- (var3);
  \draw[-]             (con13) -- (var1);

  \draw[-]             (con35) -- (var3);
  \draw[-]             (con35) -- (var5);

  \draw[-]             (con56) -- (var5);
  \draw[-]             (con56) -- (var6);
\end{tikzpicture}  
\subcaption{Output variable scores}
\end{subfigure}
\begin{subfigure}[c]{0.49\columnwidth}
\begin{tikzpicture}[node distance=1.2cm,
    every node/.style={fill=white, font=\small}, align=center]
  \node (var1)     [gnn_var, draw=red,fill=red]          {$$};
  \node (var2)     [right of=var1, xshift=3em,  gnn_var,draw=red, fill=red]          {$$};

  \node (var3)     [below of=var1, gnn_var]          {$$};
  \node (var4)     [below of=var2, gnn_var]          {$$};
  \node (var5)     [below of=var3, gnn_var, draw=red,fill=red]          {$$};
  \node (var6)     [below of=var4, gnn_var, draw=red, fill=red]          {$$};

  \node (con12)     [right of = var1 , xshift=-.5em,gnn_con, draw=red, fill=red]          {$$};
  \node (con13)     [below of = var1 , yshift=2em, xshift=-2em, gnn_con]          {$$};
  \node (con35)     [below of = var3 , yshift=2em, xshift=-2em, gnn_con]          {$$};
  \node (con234)     [below of = con12 ,yshift=1.5em, gnn_con]          {$$};

  \node (con546)     [below of = con234 , yshift=1em, gnn_con]          {$$};

  \node (con56)     [right of = var5 , gnn_con, color=red]          {$$};

  \draw[-,color=red]             (con12) -- (var1);
  \draw[-,color=red]             (con12) -- (var2);

  \draw[-]             (con234) -- (var3);
  \draw[-]             (con234) -- (var4);
  \draw[-]             (con234) -- (var2);

  \draw[-]             (con546) -- (var5);
  \draw[-]             (con546) -- (var4);
  \draw[-]             (con546) -- (var6);

  \draw[-]             (con13) -- (var3);
  \draw[-]             (con13) -- (var1);

  \draw[-]             (con35) -- (var3);
  \draw[-]             (con35) -- (var5);

  \draw[-,color=red]             (con56) -- (var5);
  \draw[-,color=red]             (con56) -- (var6);

\end{tikzpicture}

\subcaption{Infeasible subgraphs}
\end{subfigure}


\caption{Overview of our approach to detect minimal infeasible subgraphs in a Factored-NLP. \textit{(a)} The input of the model is a Factored-NLP. Circles represent variables and squares are constraints. \textit{(b)} We perform several iterations of neural message passing using the structure of the NLP. \textit{(c)} The network outputs the probability that a variable belongs to a minimal infeasible subgraph. \textit{(d)} We extract several minimal infeasible subgraphs with a connected component analysis.} 
  \label{fig:overview}
\vspace{-0.5cm}
\end{figure}
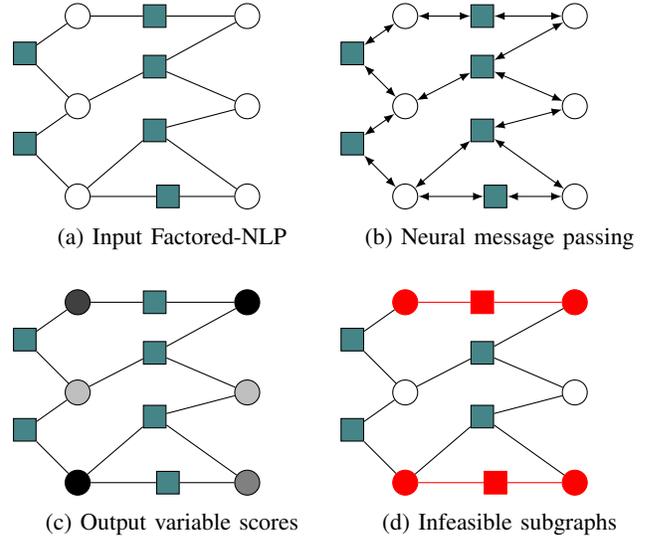

In continuous domains, extracting conflicts requires solving multiple nonlinear programs (NLPs) adding and removing constraints. While the number of NLPs required to solve grows logarithmically with the number of nonlinear constraints, each call to a nonlinear solver is usually expensive.



In this paper, we propose a neural model to predict minimal infeasible subsets of variables and constraints from a factored nonlinear program. 
The input to our model is directly the factored nonlinear program, including semantic information on variables and constraints (e.g. a class label) and a continuous feature for each variable (which, for instance can be used to encode the geometry of a scene in robotics).
The graph structure of the factored nonlinear program is exploited for performing message passing in the neural network.
Finding the minimal infeasible subgraph (i.e. a subset of variables and constraints of the Factored-NLP) is cast as a variable classification problem, and the predicted infeasible subsets are extracted with a connected components analysis.
An overview of our approach is shown in Fig. \ref{fig:overview}.
The prediction of the graph neural network can be naturally integrated into an algorithm to detect minimal infeasible subgraphs, providing a significant speedup with respect to previous methods, both using general conflict extraction algorithms or expert algorithms with domain knowledge.

Our approach follows a promising trend in robotics to combine optimization and learning
\cite{
20-driess-ICRA, driess2020deep,mansard2018using, cauligi2020learning, deits2019lvis}. In this paradigm, a dataset of solutions to similar problems is used to accelerate optimization methods, making expensive computations tractable and enabling real-time solutions to combinatorial and large scale optimization. 
Therefore, we assume that a dataset of labeled factored nonlinear programs (generated offline with an algorithm for conflict extraction) is available.
Each example consists of a factored nonlinear program and a set of minimal infeasible subgraphs.









As an application, we evaluate our method in robotic sequential manipulation.  Finding minimal conflicts is a fundamental step in conflict-based solvers, usually accounting for most of the computational time.



From a robotics perspective, our novel contribution is to use the structure of the nonlinear program formulation of manipulation planning for message passing with a graph neural network. To this end, we first formulate the motion planning problem that arises from a high-level manipulation sequence (symbolic actions such as pick or place) as a factored nonlinear program \cite{toussaint2018differentiable}.  Variables correspond to the configuration of objects and robots at each step of the motion and the nonlinear constraints model kinematics, collision avoidance and grasping constraints.
When combined with our learned model, we get strong generalization capabilities to predict the minimal infeasibility of manipulation sequences of different lengths 
in different scenes, involving a different number of objects and robots.

Our contributions are:
\begin{itemize}
  \item A neural model to predict minimal conflicts in factored nonlinear programs. We formulate the detection of minimal infeasible subgraphs as a variable classification problem and a connected components analysis.
  \item An algorithm that integrates the prediction of our neural model and non-learning conflict extraction methods, providing a significant acceleration.
  \item 
    An empirical demonstration that the formulation of manipulation planning as a factored nonlinear program, together with our neural model, enables scalability and generalization. 
\end{itemize}












\section{Related Work}

\subsection{Minimal Infeasible Subsets of Constraints}

In the discrete SAT and CSP literature, a minimal infeasible subset of constraints (also called Minimal Unsatisfiable Subset of Constraints or Minimal Unsatisfiable Core)
is usually computed by solving  a sequence of SAT and MAX-SAT problems \cite{liffiton2008algorithms,marques2021conflict, hemery2006extracting}.

In a general continuous domain, a minimal infeasible subset can be found by solving a linear number of problems  \cite{amaldi1999some}. This search can be accelerated with a divide and conquer strategy, with logarithmic complexity  \cite{junker2004preferred}. In convex and nonlinear optimization, 
we can find approximate minimal subsets by solving one optimization program with slack variables \cite{shoukry2018smc}.
In contrast, our method uses learning to directly predict minimal infeasible subsets of variables and constraints, and can be combined with these previous approaches to reduce the computational time.

\subsection{Graph Neural Networks in Combinatorial Optimization}

We use Graph Neural Networks (GNN) \cite{kipf2016semi, battaglia2018relational, ma_tang_2021} for learning in graph-structured data. 
Different message passing and convolutions have been proposed, e.g. \cite{gilmer2017neural,velivckovic2017graph}. Our architecture, targeted towards inference in factored nonlinear programs, is inspired by previous works that approximate belief propagation in factor graphs \cite{zhang2020factor, garcia2020neural, kuck2020belief}.

Recently, GNN models have been applied to solve NP-hard problems \cite{schuetz2021combinatorial}, Boolean Satisfaction \cite{selsam2018learning}, Max cut  \cite{yao2019experimental}, constraint satisfaction \cite{toenshoff2021graph}, and discrete planning \cite{shen2020learning,rivlin2020generalized,DBLP:conf/socs/NirSK21}. 
Compared to state-of-the-art solvers, learned models achieve competitive solution times and scalability but are outperformed in reliability and accuracy.
To our knowledge, this is the first work to use a GNN model to predict the minimal infeasible subgraphs of a factored nonlinear program in a continuous domain. 







\subsection{Graph Neural Networks in Manipulation Planning}

In robotic manipulation planning, GNNs are a popular architecture to represent the relations between movable objects, because they provide a strong relational bias and a natural generalization to including additional objects in the scene. 

For example, they have been used as problem encoding to learn policies for robotic assembly \cite{pmlr-v164-funk22a,Ghasemipour2022blocks} and manipulation planning \cite{li2020towards}, to learn object importance and guide task and motion planning \cite{silver2021planning}, and to learn 
dynamical models and interactions between objects \cite{driess2022learning}, \cite{paus2020predicting}.
Previous works often use object centric representations: the vertices of the graph represent the objects and the task is encoded in the initial feature vector of each variable.
Alternatively, our model performs message passing using the structure of the nonlinear program of the manipulation sequence, achieving generalization to different manipulation sequences that fulfil different goals.


\section{Formulation}

\subsection{Factored Nonlinear Program (Factored-NLP)}

A Factored-NLP $G$ is a bipartite graph $G=(X \cup H,E)$
that models the dependencies between a set of variables $ X = \{x_i \in \R^{n_i} \}$ and a set of constraints
$H = \{ h_a : \R^{m_a} \to \R^{m_a'} \}$. Each constraint $h_a(x_a)$ is a piecewise differentiable function 
evaluated on a (typically small) subset of variables  $x_a \subseteq X $  (e.g. $x_a = \{x_1, x_4\}$).
The edges model the dependencies between variables and constraints $E = \{(x_i, h_a): \text{constraint $h_a$ depends on variable $x_i$} \}$. Throughout this document, we will use the indices $i,j$ to denote variables and $a,b$ to denote constraints.

The underlying/associated nonlinear program is
\begin{equation} 
\label{eq:feasibility}
  \text{find} ~ x_i  ~ \text{s.t.} ~  h_a(x_a) \leq 0  \quad \forall x_i \in X ,h_a \in H~.
\end{equation}
The constraints $ h_a(x_a) \leq 0$ also include equality constraints (that can be written as $h_a(x_a) \leq 0$ and $h_a(x_a) \geq 0$).

A Factored-NLP $G$ is feasible ($\mathcal{F}(G) = 1$) iff \eqref{eq:feasibility} has a solution, that is, there exists a value assignment $\bar{x}_i$ for each variable $x_i$ such that all constraints $ h_a( \bar{x}_a)$  are fulfilled. 
Otherwise, it is infeasible ($\mathcal{F}(G)=0$). This assignment can be computed with nonlinear optimization methods, such as Augmented Lagrangian or Interior Points.
A minimal infeasible subgraph $M\subseteq G$ is an infeasible subset of variables and constraints  whose any proper subset is feasible,
\begin{equation}
M \subseteq G , ~\mathcal{F}(M)=0, ~ \mathcal{F}(M')= 1 ~ \forall M' \subset M.
\end{equation}




Given a graph $G$ and a subset of variables $X' \subseteq X$, a \emph{variable-induced} subgraph  $G[X'] = (X'\cup H', E')$ with $H' = \{ h_a \in H : \text{Neigh}_G(h_a) \subseteq X' \}$ is the subgraph spanned by the variables $X'$. Intuitively, $G[X']$ contains the variables $X'$ and all the constraints that can be evaluated with these variables. In this work, we consider only minimal subgraphs in the form of \textit{variable-induced} subgraphs, i.e. $M = G[X'] \subseteq G$, 
because it enables a more compact representation 
(our approach can be adapted to predict general subgraphs if required, changing the proposed variable classification to constraint classification in Sec \ref{sec:asnodeclassif}).

%

A minimal infeasible subgraph is connected and a supergraph $\tilde{M} \supseteq M$ of an infeasible subgraph $M$ is also infeasible. A factored-NLP 
$G$ can contain multiple infeasible subgraphs,
and a variable $x_i \in X$ can belong to multiple infeasible subgraphs.


\subsection{Minimal Infeasible Subgraph as Variable Classification }
\label{sec:asnodeclassif}



Let $\Phi_G = \{ M_r \subseteq G \}$ be the set of minimal infeasible subgraphs of a Factored-NLP $G$. 
Instead of learning the mapping $\phi : G \to \Phi_G$ directly, we propose to learn an over-approximation $\tilde{\phi}$
that can efficiently be framed as binary variable classification. 

We first introduce the \textit{variable-feasibility} function $\psi(x_i;G)$
that assigns a label $y_i \in \{0,1\}$ to each  variable $x_i \in X$.  $y_i=0$ if $x_i$ belongs to some infeasible subgraph $M_r$, and $y_i=1$ otherwise.
Given such a labelled graph, we can recover the infeasible subgraphs approximately by computing the connected components on the subgraph induced by the variables with label $0$, i.e. $ G\left[\{ x_i \in X : y_i = 0 \} \right]$. Thus, we define the approximate mapping as,
\begin{equation}
\tilde{\phi}(G) = \text{CCA}\left(G\left[\{ x_i \in X: y_i = 0 \}\right] \right),
\end{equation}
where $\text{CCA}$ denotes a connected component analysis.


The approximate mapping $\tilde{\phi}$ is exact, i.e. $\tilde{\phi} = \phi$, if the infeasible subgraphs are  disconnected. If two or more of the infeasible subgraphs are connected, it returns their union as a minimal infeasible sugraph, i.e. $\cup \tilde{\phi} = \cup \phi$, which over-approximates the size of the original minimal infeasible subgraph. 
Our neural model will be trained to imitate the labels of the \textit{variable-feasibility} function $\psi$. 

We emphasize that learning the approximate function $\tilde{\phi}$ is not a real limitation. First, because the prediction will be integrated into an algorithm that can further reduce the size of the infeasible subgraph, if it is not already minimal, as shown later in Sec.~\ref{sec:algorithm}. Second, because in practice finding small infeasible subgraphs, as opposed to strictly minimal, is already useful in the applications.
Finally, note that $\phi$ can be converted to a multiclass variable classification $f(x_i;G) = y \subseteq \{1,\ldots,R\}$, where each variable can belong to 
multiple classes -- but this would require a complex, and potentially intractable, permutation invariant formulation.



\subsection{GNN with the Structure of a Factored-NLP}

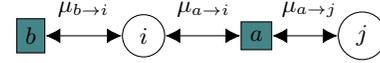
\begin{figure}
	\centering
\begin{tikzpicture}[node distance=1.5cm,
    every node/.style={fill=white, font=\small}, align=center]
  \node (vari)     [gnn_var]          {$i$};
  \node (varj)     [right of=vari, xshift=4em,  gnn_var]          {$j$};



  \node (cona)     [right of = vari , gnn_con]          {$a$};
  \node (conb)     [left of = vari,  gnn_con]          {$b$};




  \draw[<->]             (cona) -- (vari);
  \draw[<->]              (cona) -- (varj) ;

  \draw[<->]             (conb) -- (vari);





  \node (mu1)     [right of = vari , xshift = -2em, yshift=1em,  sec]          {$\mu_{a \to i}$};
  \node (mu2)     [right of = vari , xshift = +2em, yshift=1em, sec]          {$\mu_{a\to j}$};

  \node (mu3)     [left of = vari , xshift = +2em, yshift=1em, sec]          {$\mu_{b\to i}$};

  \node (planner)     [process,below of=vari,yshift=1em, xshift=1cm]          {
      $\left[ \mu_{a \to i} , \mu_{a \to j } \right] = \text{Message}_a(  z_i , z_j  )$,~~
      $\mu_{b \to i}  = \text{Message}_b(  z_i   ) $  \\
      $z_i' = \text{Update}\left( \text{AGG}\left( \mu_{a\to i } , \mu_{b \to i } \right) , z_i \right) $,~~
      $z_j' = \text{Update}\left( \mu_{ a \to j}  , z_j \right) $ 
};

\end{tikzpicture}

\caption{Message Passing in a Factored-NLP with two variables ($i,j$) and two constraints $(a,b$).}
  \label{fig:message_pass}
  \vspace{-0.5cm}
\end{figure}

A fundamental idea of our method is to use the structure of the Factored-NLP for message passing with Graph Neural Networks (GNN) to learn the \emph{variable-feasibility} $\psi(x_i;G)$. 

Each variable vertex $x_i \in X$ has a feature vector $z_i \in \R^{n_z}$ that is updated with the incoming messages of the neighbour constraints. $z_i$ is initialized with $z_i^0$ to encode semantic and continuous information of the variable $x_i$ (an example on how to initialize the features in manipulation planning is shown in Sec. \ref{sec:encoding}).
The update rule follows a two-step process:
first,  each constraint computes and sends back a message to each neighbour variable, which depends on the current features of all the neighbour variables. Second, each variable aggregates the information of the incoming messages from the constraints and updates its feature vector.
A graphical representation is shown in Fig \ref{fig:message_pass}.
\begin{align}
\label{eq:message_pass}
[ \oplus \mu_{a \to i} ]_{i \in N(a)} = \text{Message}_{a}( [ \oplus z_{i} ]_{i \in N(a)}),  \\
  z_i' = \text{Update} ( \text{AGG}_{a \in N(i)} ~ \mu_{a \to i} , z_i  ). \nonumber
\end{align}
 $\mu_{a\to i } \in \R^{n_{\mu}}$ is the message from constraint $a$ to variable $i$.
 $[\oplus z_i]_{i}$ denotes concatenation. $N(a) = \text{Neigh}_G(h_a)$ is the ordered set of variables connected to the constraint $h_a$. $N(i) = \text{Neigh}_G(x_i)$ is the set of constraints connected to variable $x_i$. AGG is an aggregation function, e.g. max, sum, mean or weighted average. We use max (element-wise) in our implementation. 
\texttt{Update} and $\texttt{Message}_a$ are small MLPs (Multilayer Perceptron) with learnable parameters. 
Likewise the nonlinear constraints in the Factored-NLP are not permutation invariant or symmetric, 
the features $z_i$ have to be concatenated in a predefined order $N(a)$ when evaluating $\texttt{Message}_a$.




The function \texttt{Update} is shared by all vertices (which generalizes to Factored-NLPs with additional variables). 
The function $\texttt{Message}_a$ is shared between different constraints of the Factored-NLP that represent the same mathematical function, i.e. 
$\texttt{Message}_a = \texttt{Message}_b ~ \text{iff} ~ h_a(x) = h_b(x)$
(which generalizes to Factored-NLPs with additional constraints).
For example, in manipulation planning, all constraints that model collisions between objects will share the same $\texttt{Message}$ MLP.

The message passing update \eqref{eq:message_pass} is performed $K$ times, starting from the initial feature vectors $z_i^0$.
The feature vectors after $K$ iterations are used for feasibility prediction with a small MLP classifier.
\begin{equation}
  \hat{y}_i = \text{Classifier}(z_i^K)
  \label{eq:node_classif}
\end{equation}
 The parameters of the classifier, message and update networks 
 are trained end-to-end to minimize the weighted binary cross entropy loss between the prediction $\hat{y}_i$ and the \textit{variable-feasibility} labels $y_i$.

\subsection{Algorithm to Detect Minimal Infeasible Subgraphs}
\label{sec:algorithm}


To account for the approximation of our variable classification formulation, and small prediction errors, we can integrate the learned classifier into a classical algorithm to detect minimal infeasible subgraphs. 

We assume the user provides the $\texttt{Solve}$ and $\texttt{Reduce}$ routines, that respectively check if a Factored-NLP is feasible and compute a minimal infeasible subset of an infeasible graph. $\texttt{Reduce}$ is an expensive routine, as it requires solving several nonlinear programs adding and removing constraints. The number of evaluated NLPs (and therefore the computation time) depends on the size of the input graph: linear on the total number of variables using \cite{amaldi1999some}, or logarithmic \cite{junker2004preferred}.







Our algorithm is shown in Alg. \ref{alg:overview}. 
The GNN model is evaluated on the input Factored-NLP and computes a feasibility scores $\hat{y_i}$ for each variable.
Iteratively increasing the classification threshold $\delta$,  we select the candidate infeasible variables $X_{\delta}$ using the current threshold $\delta$. We generate candidate infeasible subgraphs 
with a connected component analysis on the \textit{variable-induced} subgraph $G[X_{\delta}]$,
that are evaluated with $\texttt{Solve}$. Once an infeasible subgraph is found, we use $\texttt{Reduce}$ to get a minimal infeasible subgraph.

A non-learning approach runs \texttt{solve} and  \texttt{reduce} directly on the input Factored-NLP. Therefore, the acceleration in our algorithm comes from evaluating these routines with small (ideally minimal) candidates.
Alg. \ref{alg:overview} can be extended to compute several minimal infeasible subgraphs by removing the break instruction (line \ref{lbl:break}) and adding a special check before solving a candidate subgraph (line \ref{lbl:solve}), to avoid solving for a supergraph of a found infeasible subgraph.

\begin{algorithm}[t]
    \caption{Conflict Extraction with a GNN}
    \label{alg:overview}
    \DontPrintSemicolon

    \SetKwFunction{AddPrimitives}{AddPrimitives}
    \SetKwFunction{ExtractPrimitives}{ExtractPrimitives}
    \SetKwFunction{ComputeDelta}{ComputeDelta}
    \SetKwFunction{DiscontinuityBoundedAstar}{db-A*}
    \SetKwFunction{Optimization}{Optimization}
    \SetKwFunction{Convert}{Convert}
    \SetKwFunction{Report}{Report}
    \SetKwFunction{Threshold}{Threshold\_select}
    \SetKwFunction{Connected}{Connected\_components}
    \SetKwFunction{Solve}{Solve}
    \SetKwFunction{Reduce}{Reduce}
    \SetKwFunction{GraphNLP}{Factored-NLP}
    \SetKwFunction{Model}{GNN\_Model}
    \SetKw{break}{break}


    \KwData{\GraphNLP $G$ , 
     \Model, 
  \Solve,
    \Reduce
 } 

  \KwResult{$M \subseteq G$ \quad  \Comment{Minimal infeasible subgraph} }

$\{\hat{y}_i\} = \Model (G)$ 

$\delta \leftarrow  0.5, ~ \delta_r \leftarrow 1.2$ , \texttt{found} $ \leftarrow 0 $ \;
\While{\textup{not \texttt{found}} }{
      $ X_{\delta} = \{ x_i \in X :  \hat{y}_i < \delta\} \quad$  \Comment{Candidate infeasible variables}
      \For{$ g \in \Connected (G[X_{\delta}])$}{
        $ \texttt{feasible} \leftarrow  \Solve ( g ) $ \; \label{lbl:solve}
        \If{\textup{not \texttt{feasible}}} {
        $ M \leftarrow \Reduce( g)    $ \;
        $ \texttt{found} \leftarrow 1  $ \;
        \break \label{lbl:break}
      }

    }

        $\delta \leftarrow \delta \times \delta_r \;$

  }
  \Return{$M$}

\end{algorithm}

\section{Factored-NLP for Manipulation Planning}

\begin{figure}
  \centering
\begin{tikzpicture}[scale=0.7,every node/.style={transform shape}]

        \node[latent] (a0) {$a_0$} ;


        \node[latent,below=.5 of a0  ] (b0) {$b_0$} ;

        \node[latent, ,below=.5 of b0] (A0) {$A_0$} ;

        \node[latent,below=.5 of A0  ] (B0) {$B_0$} ;

        \node[latent,below=.5  of B0] (q0) {$q_0$} ;
        \node[latent,below=.5  of q0] (w0) {$w_0$} ;




        \node[latent,right=2 of a0  ] (a1) {$a_1$} ;
        \node[latent,below=.5 of a1  ] (b1) {$b_1$} ;

        \node[latent, ,below=.5 of b1] (A1) {$A_1$} ;

        \node[latent,below=.5 of A1  ] (B1) {$B_1$} ;

        \node[latent,below=.5 of B1] (q1) {$q_1$} ;
        \node[latent,below=.5 of q1] (w1) {$w_1$} ;


        \node[latent,right=2 of a1  ] (a2) {$a_2$} ;
        \node[latent,below=.5 of a2  ] (b2) {$b_2$} ;
        \node[latent,below=.5 of b2  ] (A2) {$A_2$} ;
        \node[latent,below=.5 of A2  ] (B2) {$B_2$} ;
        \node[latent,below=.5 of B2] (q2) {$q_2$} ;
        \node[latent,below=.5 of q2] (w2) {$w_2$} ;



        \node[latent,right=2 of a2  ] (a3) {$a_3$} ;
        \node[latent,below=.5 of a3  ] (b3) {$b_3$} ;

        \node[latent,below=.5 of b3  ] (A3) {$A_3$} ;
        \node[latent,below=.5 of A3  ] (B3) {$B_3$} ;

        \node[latent,below=.5 of B3] (q3) {$q_3$} ;
        \node[latent,below=.5 of q3] (w3) {$w_3$} ;


      \factor[left=1 of A0, yshift=0.0cm] {trajp0} { PoseDiff } {A0, a0} {};
      \factor[left=1 of B0, yshift=0.0cm] {trajp0} { PoseDiff } {B0, b0} {};
      \factor[left=1 of w0, yshift=0.5cm] {trajp0} { Ref } {w0} {};
      \factor[left=1 of q0, yshift=0.5cm] {trajp0} { Ref } {q0} {};
      \factor[left=1 of a0, yshift=-0.5cm] {trajp0} { Ref } {a0} {};
      \factor[above=.3 of a1,xshift=-.5cm] {trajp0} { left:Ref } {a1} {};
      \factor[above=.3 of a2, xshift=-.5cm] {trajp0} { left:Ref } {a2} {};

      \factor[left=1 of A1, yshift=-0.5cm] {trajp0} { PoseDiff } {A1, a1} {};

      \factor[left=1 of B1, yshift=0.0cm] {trajp0} { PoseDiff } {B1, b1, q1} {};

      \factor[left=1 of b0, yshift=0.0cm] {trajp0} { Ref } {b0} {};
      \factor[left=1 of b1, yshift=0.5cm] {trajp0} { below:Grasp } {b1} {};
      \factor[left=1 of b2, yshift=0.5cm] {trajp0} { Grasp } {b2} {};
      \factor[right=1 of b3, yshift=0.5cm] {trajp0} { Pos } {b3} {};
      \factor[above=.3 of a3,xshift=-.5cm] {trajp0} {left:Ref} {a3} {};




      \factor[left=1 of b1, yshift=-.5cm] {trajp0} {below:Kin} {b0, q1,b1} {};

      \factor[right=1.2 of b1, yshift=-.3cm] {trajp0} {Kin} {b1, q2,b2,w2} {};

      \factor[left=.7 of b3, yshift=-.3cm] {trajp0} {Kin} {w3,b3,A3,b2} {};

      \factor[left=1 of a3] {trajp0} {Equal} {a2,a3} {};

      \factor[left=1 of a1] {trajp0} {Equal} { a0, a1} {};

      \factor[left=1 of a2] {trajp0} {Equal} { a1, a2} {};

      \factor[left=1 of A2, yshift=-0.5cm] {trajp0} { PoseDiff } {A2, a2} {};
      \factor[left=1 of A3, yshift=-0.5cm] {trajp0} { PoseDiff } {A3, a3} {};
      \factor[left=1 of B2, yshift=0.0cm] {trajp0} { PoseDiff } {B2, b2, w2} {};
      \factor[left=1 of B3, yshift=0.0cm] {trajp0} { PoseDiff } {B3, b3, A3} {};

      \factor[right=.3 of A0, yshift=-0.5cm,color=brown] {} {} {A0,B0} {};
      \factor[right=.3 of B0, yshift=-0.5cm,color=brown] {} {} {B0,q0} {};
      \factor[right=.3 of q0, yshift=-0.1cm,color=brown] {} {} {A0,q0} {};

      \factor[right=.3 of q0, yshift=-0.5cm,color=brown] {} {} {q0,w0} {};
      \factor[left=.9 of w0, yshift=-0.5cm,color=brown] {} {} {A0,w0} {};
      \factor[left=.2 of w0, yshift=+1.0cm,color=brown] {} {} {w0,B0} {};

      \factor[right=.3 of A1, yshift=-0.5cm,color=brown] {} {} {A1,B1} {};
      \factor[right=.3 of B1, yshift=-0.5cm,color=brown] {} {} {B1,q1} {};
      \factor[right=.3 of q1, yshift=-0.1cm,color=brown] {} {} {A1,q1} {};

      \factor[right=.3 of q1, yshift=-0.5cm,color=brown] {} {} {q1,w1} {};
      \factor[left=.9 of w1, yshift=-0.5cm,color=brown] {} {} {A1,w1} {};
      \factor[left=.2 of w1, yshift=+1cm,color=brown] {} {} {w1,B1} {};

      \factor[right=.3 of A2, yshift=-0.5cm,color=brown] {} {} {A2,B2} {};
      \factor[right=.3 of B2, yshift=-0.5cm,color=brown] {} {} {B2,q2} {};
      \factor[right=.3 of q2, yshift=-0.1cm,color=brown] {} {} {A2,q2} {};

      \factor[right=.3 of q2, yshift=-0.5cm,color=brown] {} {} {q2,w2} {};
      \factor[left=.9 of w2, yshift=-0.5cm,color=brown] {} {} {A2,w2} {};
      \factor[left=.3 of w2, yshift=+1cm,color=brown] {} {} {w2,B2} {};

      \factor[right=.3 of A3, yshift=-0.5cm,color=brown] {} {} {A3,B3} {};
      \factor[right=.3 of q3, yshift=-0.1cm,color=brown] {} {} {A3,q3} {};

      \factor[right=.3 of q3, yshift=-0.5cm,color=brown] {} {} {q3,w3} {};
      \factor[left=.9 of w3, yshift=-0.5cm,color=brown] {} {} {A3,w3} {};
      \factor[left=.2 of w3, yshift=+1cm,color=brown] {} {} {w3,B3} {};
      \factor[right=.3 of B3, yshift=-0.5cm,color=brown] {} {} {B3,q3} {};







  \end{tikzpicture}
  \caption{Factored-NLP in manipulation planning.
    Circles are variables and squares are constraints. Each column represents a keyframe of the manipulation sequence. $q,w $ are the configurations of two robots; $A,B$ are the absolute position of two objects, and $a,b$ are the relative pose of these objects with respect to their parent in the kinematic tree (e.g. the table, a gripper or another object). See main text for details.
}
  \label{fig:cg_example2}
  \vspace{-0.5cm}
\end{figure}
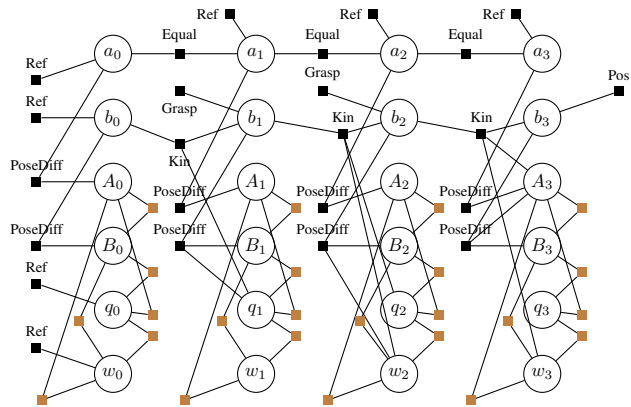

In this section, we present a factored nonlinear program formulation for robotic manipulation planning that enables our model to generalize to problems with longer manipulation sequences, more objects and robots, and different geometric environments.

\subsection{Structure of the Factored-NLP}

A Factored-NLP models the motion of robots and objects that is implied by a sequence of high-level actions (such as pick and place) as a nonlinear optimization/feasibility problem. It contains variables that represent the configuration of objects and robots at each time step.  We focus on the keyframe or mode-switch problem, that considers only the configurations at the beginning and end of each motion phase  (e.g. when picking or placing an object), but not the trajectory between them. This follows a common problem decomposition used in Task and Motion Planning, where path feasibility is evaluated afterwards with trajectory optimization or sampling based motion planning \cite{garrett2018sampling, toussaint2018differentiable}.

We include three types of variables: robot configurations, object absolute positions and object relative positions with respect to the parent frame. Nonlinear constraints model grasping (\textit{Grasp} in Fig. \ref{fig:cg_example2}), kinematic switches (\textit{Kin}), valid placements (\textit{Pos}), collision avoidance (brown squares), reference position (\textit{Ref}), time consistency (\textit{Equal}), and geometric consistency between relative and absolute pose (\textit{PoseDiff}).
The structure of the NLP is similar to previous factored formulations \cite{Ortiz2022Conflict-Directed, garrett2018sampling, lagriffoul2014efficiently, ortiz2021learning} but 
less compact (one can formulate the nonlinear program without explicitly introducing the absolute position of objects). However, this is necessary to formulate Factored-NLPs of diverse manipulation sequences using only few types
of nonlinear constraints. Because each type of constraint will correspond to a different $\texttt{Message}$ network, this formulation is crucial to enable generalization of the GNN model. 
In Fig. \ref{fig:cg_example2}, we show the Factored-NLP that corresponds to the sequence \textit{ (pick object B with robot Q), (pick object B with robot W from robot Q), (place object B on top of object A  with Robot W)}.

\subsection{Encoding of the Problem in the Initial Feature Vectors}\label{sec:encoding}


The structure of the Factored-NLP implicitly encodes the number of objects, robots and the high-level action sequence (e.g. which robots pick which objects).
The geometric description of the environment is encoded locally in the initial feature vector of each variable $z_i^0$. 
Specifically, the initial feature vector includes the information of unary constraints (i.e. constraints evaluated only on a single variable, which are then not added to the message passing architecture), additional semantic class information (for example, whether the variable represents an object or a robot, but without including a notion of time index or entity), and geometric information that is relevant for the constraints (for example, the size of the objects). The dimension of $z_i^0$ is fixed, and shorter feature vectors are padded with zeros.

For example, suppose that the Factored-NLP of Fig. \ref{fig:cg_example2} is evaluated in a scene where robot $Q$ is at pose { \small $T_Q =$ $[ \,0.32 , \,0.41 , \,0.56, \,0.707 , \, 0 ,\, 0 ,\, 0.707 ] $ }, the start  position of object $A$ is{ \small $T_A = [ 0.35, \, 0.4, \, 0.5 , \, 0.707 , \,  0 , \, 0 , \, 0.707  ]$ } and object $A$ is a box of size { \small $S_A = [0.2 , \, 0.3 , \, 0.2 ]$}. Then $z^0$ of variables $\{q_0,q_1,q_2,q_3\}$ is { \small $[ 1,\, 0,\,0,\,0,\,0,\,0 ,  T_Q  ]$ } where the first 6 components are a one-hot vector to indicate that it is a robot.
The $z^0$ of $\{a_0,a_1,a_2,a_3\}$ is  {\small $  [0, \, 1,\, 0,\, 0, \, 0, \,0 ,  \, T_A ]$ }, where first components indicate that it is a relative pose with respect to the reference position. The $z^0$ of $\{A_0,A_1,A_2,A_3\}$ is { \small 
$[  0, \, 0, \, 1, \, 0,\, 0, \, 0 , \,  S_A ,\, 0 ,\, 0 , \, 0, \, 0 ]$ } to indicate that it is an absolute position of an object of size $S_A$.

\section{Experimental Results}

\begin{figure}

\setlength{\tabcolsep}{0.2em} 
\centering
\begin{tabular}{ccc}

  \includegraphics[width=.30\columnwidth]{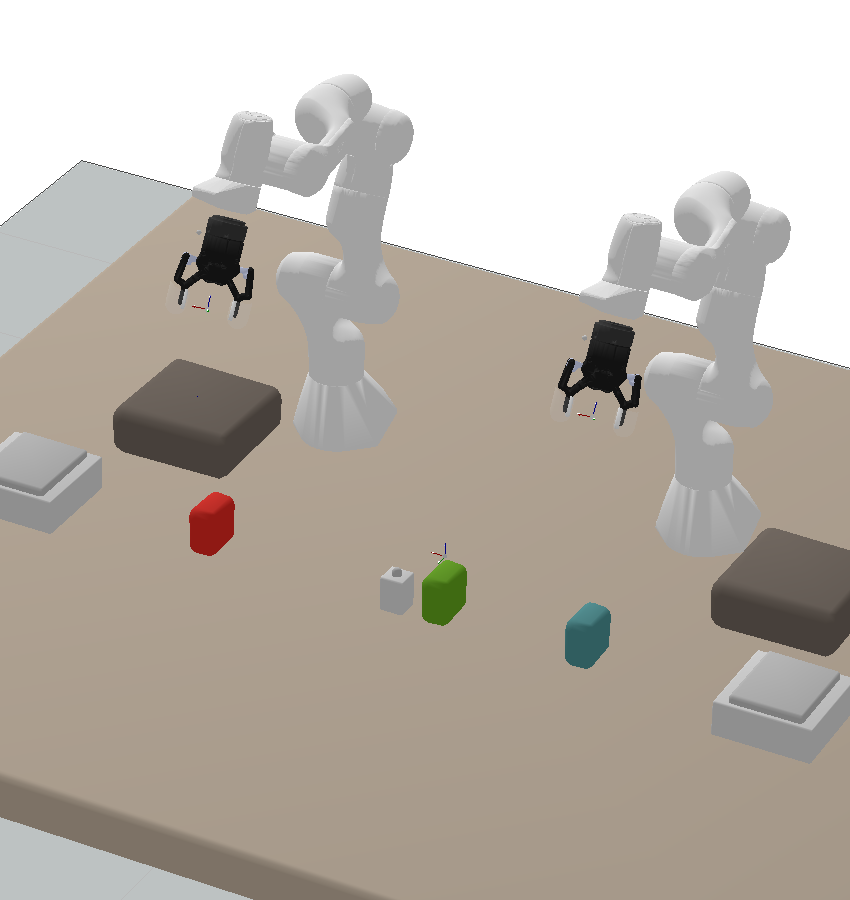} &
  \includegraphics[width=.30\columnwidth]{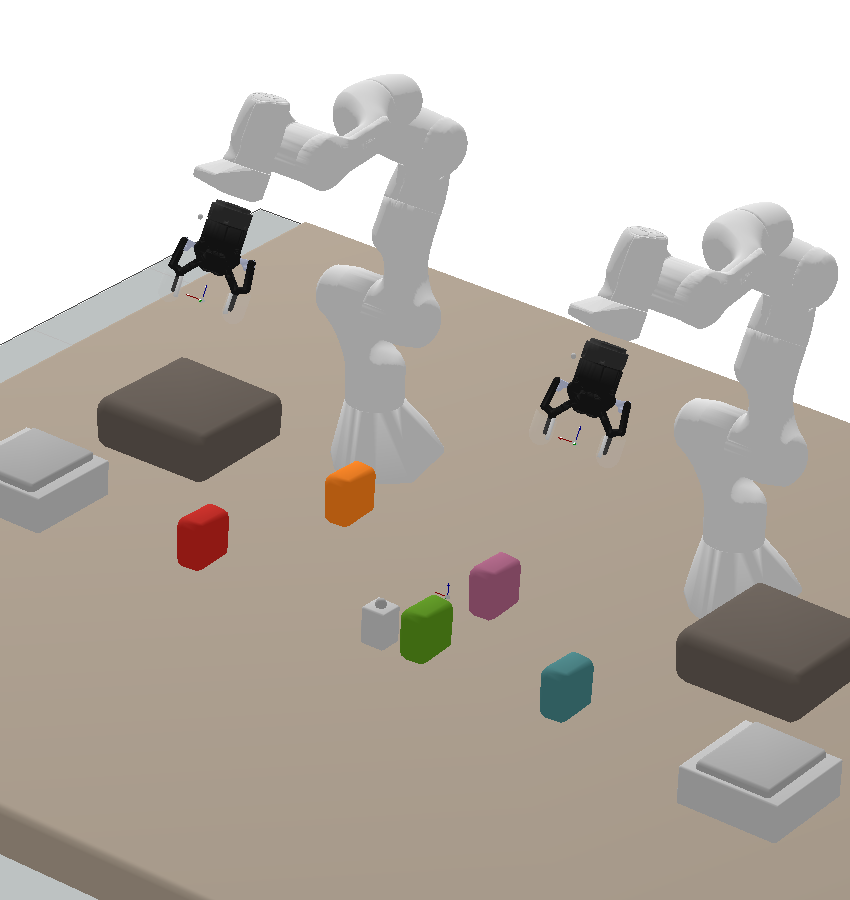} &
  \includegraphics[width=.30\columnwidth]{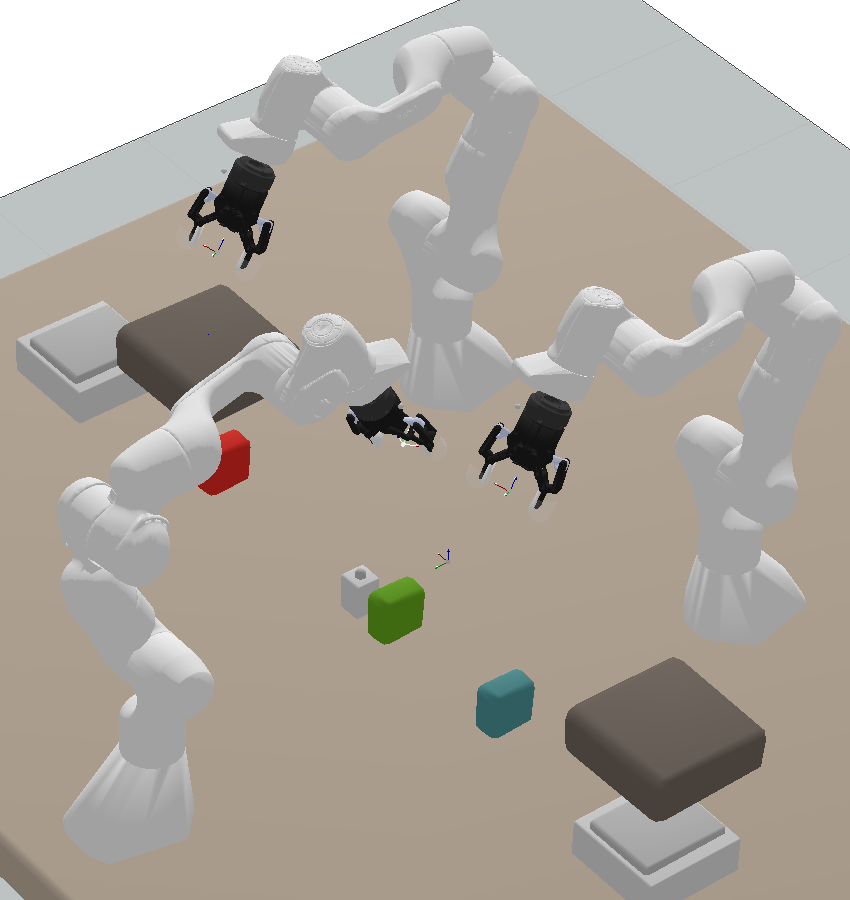}
  \end{tabular}
  \caption{Manipulation Scenarios. Obstacles are brown, blocks are colorful and 
    tables are white. \textit{Left}: Train Data, \textit{Middle}: + Blocks, \textit{Right:} + Robots.}
  \label{fig:view_data}
  \vspace{-0.6cm}
\end{figure}

\subsection{Scenario}

We evaluate our model in robotic sequential manipulation. The high-level goal is to build towers and rearrange  blocks in different configurations, in scenarios that contain several and varying number of blocks, robots and movable obstacles, at different positions, see Fig. \ref{fig:view_data} and \ref{fig:view_skeleton}. The following setting is used to generate the train dataset (4800 Factored-NLPs):

$\bullet$ Five movable objects: 3 blocks and 2 obstacles. 
    Both types of objects have collisions constraints, but obstacles are bigger and usually block grasps or placements. 
  $\bullet$ Two robots: 7-DOF Panda robot arms, that can pick and place objects using a top grasp.
 $\bullet$ Different geometric scenes: the position of the objects, robots and tables is randomized.
  $\bullet$
  Manipulation sequences of length 4 to 7.



To evaluate the generalization capabilities of the learned model, we consider three additional datasets:

$ \bullet$ \texttt{+Robots}:  we add an additional robot.
 $ \bullet $ \texttt{+Blocks}:  we add two additional blocks.
 $ \bullet $ \texttt{+Actions}:  it contains Factored-NLPs with longer manipulation sequences (length of 8 to 10).


\subsection{Data Generation}

For training the GNN model we need a set of Factored-NLPs with labelled variables to indicate whether they belong to a minimal infeasible subset.
First, we generate a set of interesting high-level action sequences. 
Second, we evaluate the manipulation sequences on random geometric scenes to generate Factored-NLPs (including the initial feature vectors). To compute the feasibility labels, we adapt the conflict extraction algorithm of \cite{Ortiz2022Conflict-Directed} to find up to 10 minimal infeasible subgraphs.

\subsection{Accuracy of the GNN Classifier}




\begin{table}

  \caption{Classification Accuracy. Each pair indicates the accuracy in predicting feasible and infeasible variables. }
  \label{tab-theaccuracy}
  \centering
\renewcommand{\arraystretch}{1.4} 
\setlength{\tabcolsep}{0.4em} 
\begin{tabular}{  l c c c c   } \toprule
 & Train Data  & + Blocks & + Robots & + Actions \\ 
 \midrule
    \emph{GNN} & $(94.7, ~95.4)  $ & $(    96.1, ~95.2  )$ & $(    95.7, ~95.3   )$  &    $(    94.6, ~94.1    )$   \\
    \emph{MLP} & $(    93.0,~82.2       )$ & $(   93.4,~80.8     )$ & $(   93.0,~80.8       )$  &    $(   91.0,~48.0    )$   \\
    \emph{MLP-SEQ} & $(   83.5,~88.1    )$ & $(    82.3,~88.8     )$ & $(  82.1,~88.8   )$  &    $(   74.0,~75.3      )$   \\
    \bottomrule
  \end{tabular}
\end{table}

We compare our model (\emph{GNN}) against a Multilayer Perceptron (\emph{MLP}) and a sequential model (\emph{MLP-SEQ}), trained with the same dataset.

The \emph{MLP} computes $\hat{y}_i = \text{MLP}( \tilde{z}^0_i, A, C) $. $\tilde{z}_i^0 = [ z_i^0 , t_i , e_i]$ is the feature vector of the variable we want to classify. It concatenates the feature vector $z_i^0$ used in the \emph{GNN}, with the time index $t_i$ of the variable, and a parametrization that defines the entity  $e_i$ (for instance, we represent an object with its starting pose -- one hot encodings could not generalize to more objects). $A$ is the encoding of the whole action sequence, using small vectors to encode each token, e.g.\ \{``pick", ``block1",  ``l\_gripper", ``table"\}. To account for sequences of different length, we fix a maximum length and add padding. $C$ is the scene parametrization and contains the position/shapes of all possible objects and robots.
We also evaluate \emph{MLP-SEQ}, a sequential model $ \text{MLP}( \tilde{z}^0_i, \text{SEQ} (A), C) $ that encodes the action sequence with a recurrent network (Gated Recurrent Units).

We first evaluate the accuracy of the models to predict if a variable belongs to a minimal infeasible subset, see Tab. \ref{tab-theaccuracy}. Our \emph{GNN} model outperforms the alternative architectures, both in the original
\textit{Train Data} and, specially, in the extension datasets. Our model keeps a constant $\sim$95\% success rate across all datasets, while the performance of \emph{MLP} and \emph{MLP-SEQ} drops to 48\% and 75\%.
We also evaluate the accuracy of our model to predict infeasible subgraphs, using the proposed method that combines variable classification  and connected component analysis, using the initial threshold for classification $\delta = 0.5$.
Our model outperforms \emph{MLP} and \emph{MLP-SEQ},  and 
finds between 70\% and 57\% of the infeasible subgraphs, and 30\%-50\% of the predicted subgraphs are minimal, see Tab. \ref{tab-theaccuracygraph}. Between  34\%-48\% of the predicted infeasible graphs are actually feasible (not shown in the table due to space limitation). 

\begin{table}
  \caption{Prediction of infeasible subgraphs. Each pair indicates the ratio ``found / total" and ``minimal / found".}
  \label{tab-theaccuracygraph}
  \centering
\renewcommand{\arraystretch}{1.4} 
\setlength{\tabcolsep}{0.4em} 
\begin{tabular}{  l c c c c   } \toprule
 & Train Data & + Blocks & + Robots & + Actions \\ 
 \midrule
    \emph{GNN} & $ (71.2,~54.1)  $ & $ (58.9,~33.3) $ & $  (~70.2,~55.3)  $  &    $  (~57.1,~41.9)  $   \\
        \emph{MLP} & $(58.5,~54.6)$ &  $(34.5,~53.2)$     &  $(55.2,~37.6)$   &  $(22.1,~35.5)$   \\
        \emph{MLP-SEQ} &$(65.7,~26.0)$&  $(28.6,~21.2)$ & $(61.3,~09.5)$ &    $(36.3,~11.0)$   \\
        \bottomrule
  \end{tabular}
\end{table}

\emph{MLP}, \emph{MLP-SEQ} and  \emph{GNN} have the same information to make the predictions. The Factored-NLP is a deterministic mapping of the action sequence and the geometric scene. Although a \emph{MLP} could learn this mapping, our experiments show that the representation does not emerge naturally --  confirming that a structured model yields better generalization.






\subsection{Finding Minimal Infeasible Subgraphs}

\begin{table}
  \caption{Finding one minimal infeasible subgraph. Each pair indicates the number of solved NLPs and the computational time in 100 Factored-NLPs, normalized by \emph{GNN+g1}.}
 
  \label{tab:infeas_subgraph}
\begin{center}
\renewcommand{\arraystretch}{1.4} 
\setlength{\tabcolsep}{0.4em} 
\begin{tabular}{  l c c c c   } \toprule
 & Train Data  & +   Blocks & + Robots & + Actions \\ 
 \midrule
    \emph{GNN+e} & $(1.57,~2.25)$  & $(1.44,~2.09)$   &  $(1.66,~2.14)$  & $(1.50,~2.19)$      \\
    \emph{GNN+g1} &  $(1,~1)$  &  $(1,~1)$ & $(1,~1)$ & $(1,~1)$ \\
    \emph{Oracle} &  $(0.83,~0.97)$  & $(0.62,~0.79)$ & $(0.83,~0.84)$    & $(0.71,~0.86)$ \\
    \emph{Expert} & $(3.66,~4.32)$  & $(3.13,~5.06)$   & $(4.33,~4.62)$  & $(3.33,~4.56)$     \\
    \emph{General 2} & $(3.50,~64.1)$  &
$(3.30,~163)$ & $(3.50,~66.5)$ & $(3.83,~128)$ \\
    \bottomrule
\end{tabular}
\end{center}
\end{table}

We analyze the time required to find one minimal infeasible subgraph in an infeasible Factored-NLP with  algorithms:

\begin{itemize}
  \item \textit{Oracle}, which executes a single call to $\texttt{Solve}$ and $\texttt{Reduce}$ with a minimal infeasible subgraph as input. 

  \item \textit{General \{1,2\}}, which are generic algorithms for conflict extraction: \emph{General 1} uses constraint filtering \cite{amaldi1999some}, and \emph{General 2} uses \textit{QuickXplain} \cite{junker2004preferred}.

  \item \textit{Expert} is a heuristic algorithm for conflict extraction in manipulation planning \cite{ortiz2022conflict}. It exploits the temporal structure, domain relaxations, and the convergence of the optimizer to quickly discover the conflicts.

  \item \textit{GNN+\{e,g1\}} combines the prediction of our \emph{GNN} model with either \emph{Expert} or \emph{General 1}. \emph{Expert} and \emph{General 1} are used as \texttt{Reduce} 
    in Alg.\ \ref{alg:overview}.


\end{itemize}

Results are shown in Table \ref{tab:infeas_subgraph}. 
\emph{GNN+g1} is 60-120x faster than \emph{General 2} (which is faster than \emph{General 1}). This highlights the benefits of our approach in domains where we can compute a dataset using \textit{General} offline, and train the model to get an order-of-magnitude improvement in new problems. \emph{GNN+g1} is 4-5x faster than the \textit{Expert} algorithm, and only 1.2x slower than an oracle.
 Moreover, the acceleration provided by \textit{GNN} is maintained in all the datasets. This confirms the good accuracy and generalization of the architecture seen in the classification results.
As a side note, 
\textit{Expert} is faster than \textit{General 2} because it solves a lot of small feasible NLPs first, until it finds one that is infeasible (which is faster than solving infeasible NLPs).







\begin{figure}

\setlength{\tabcolsep}{0.1em} 
\centering
\begin{tabular}{ccc}
  \includegraphics[width=.31\columnwidth]{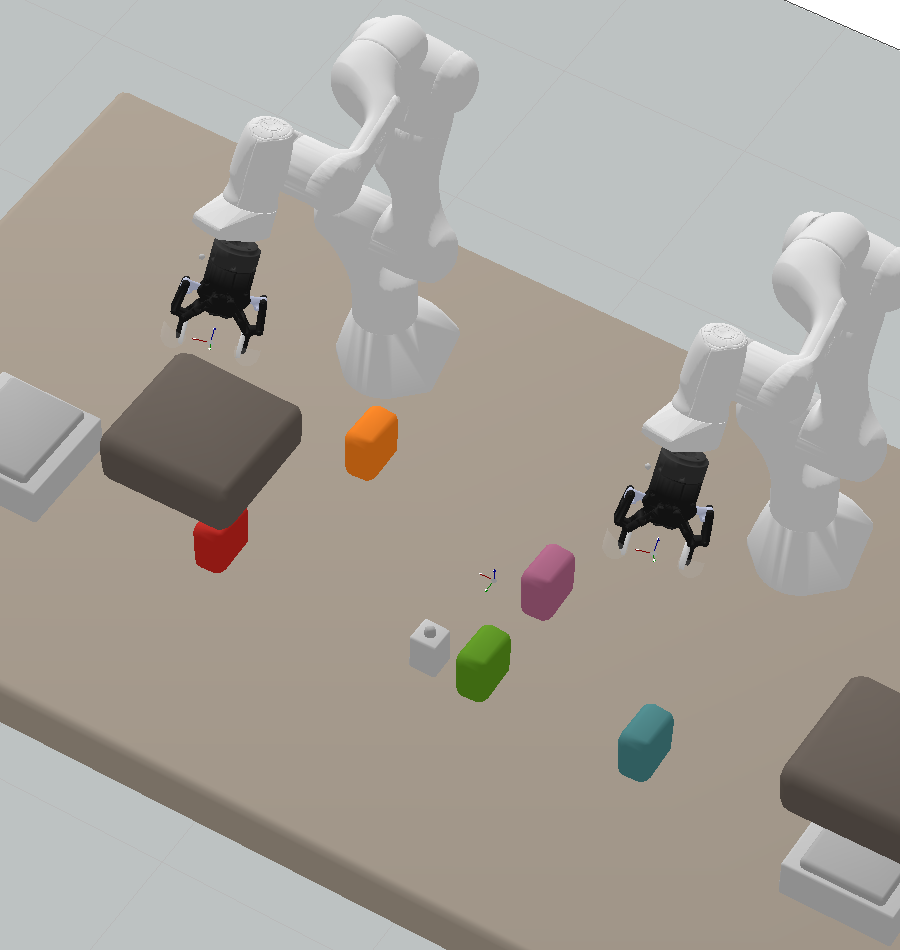} &
  \includegraphics[width=.31\columnwidth]{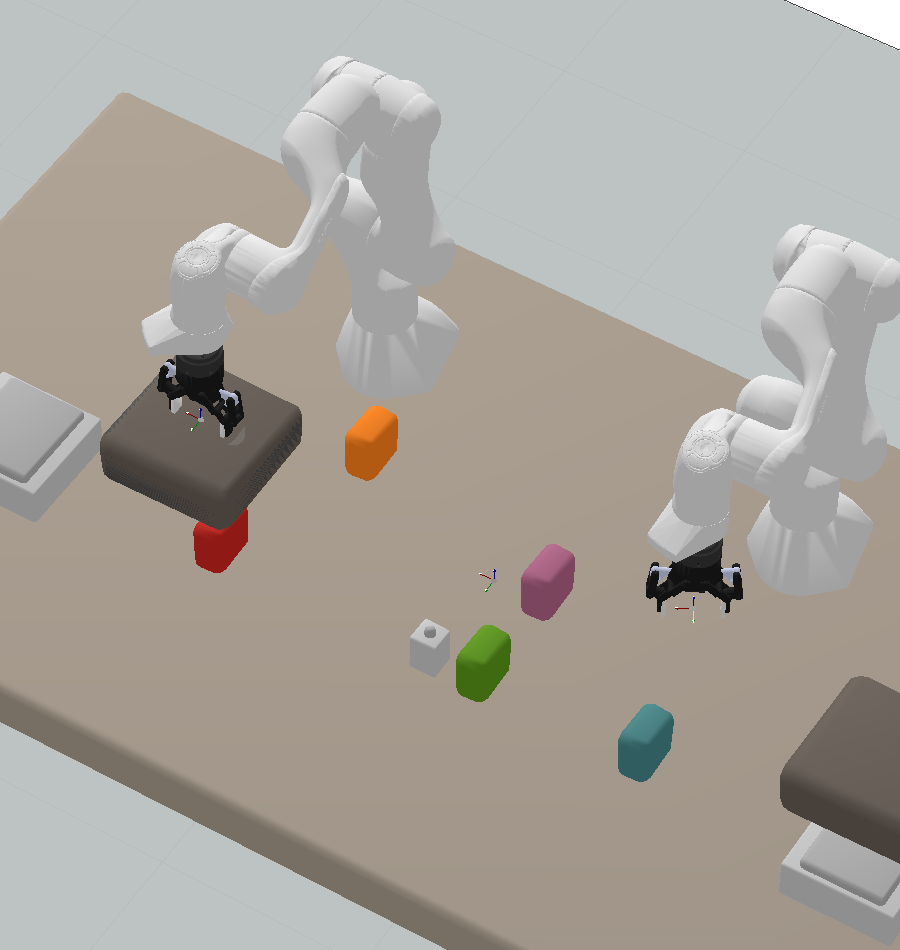} &
  \includegraphics[width=.31\columnwidth]{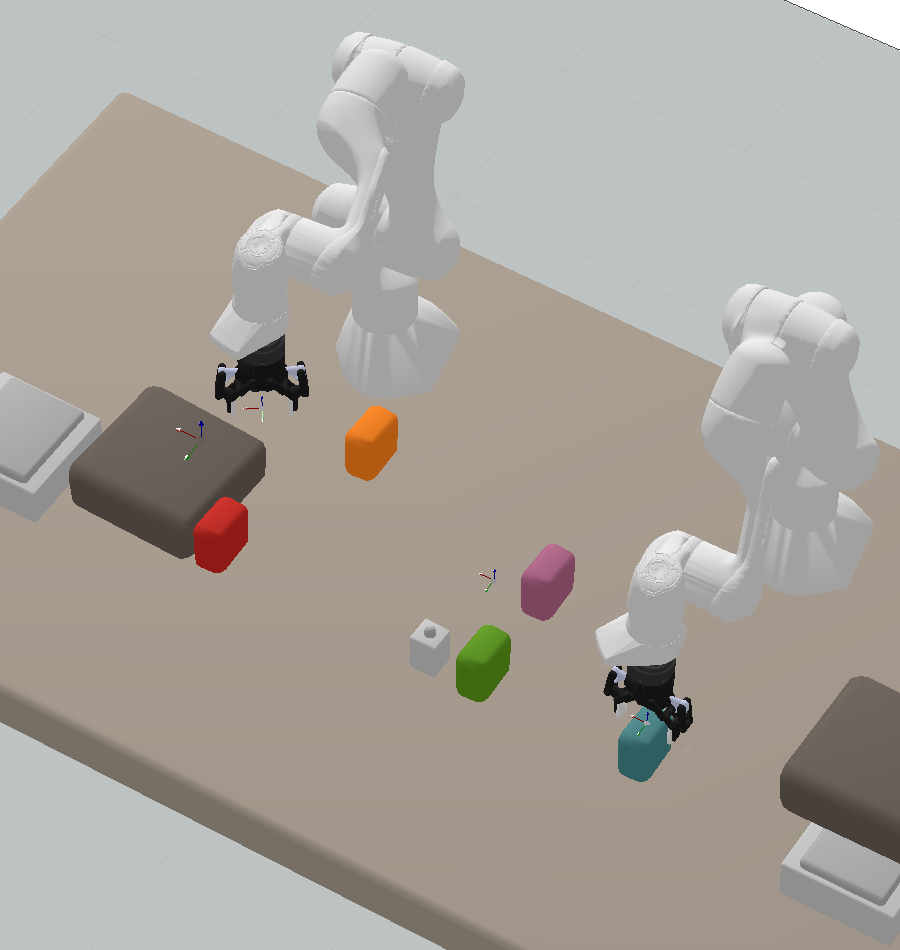} \\
  \includegraphics[width=.31\columnwidth]{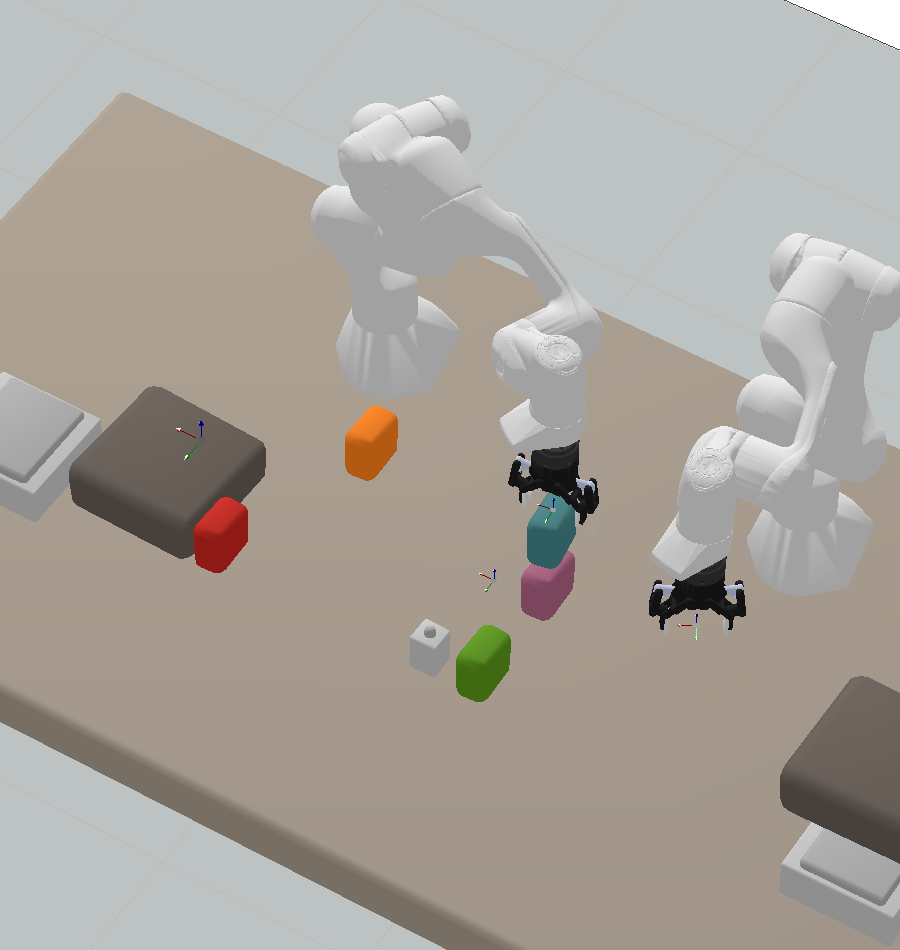} &
  \includegraphics[width=.31\columnwidth]{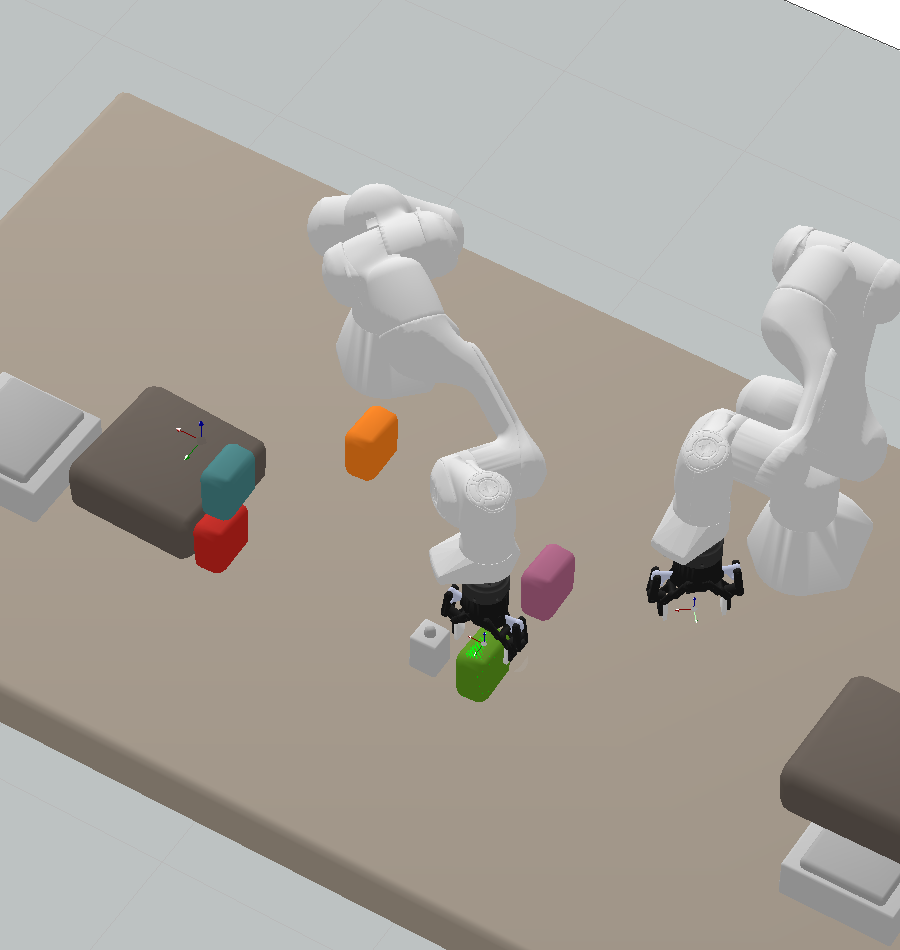} &
  \includegraphics[width=.31\columnwidth]{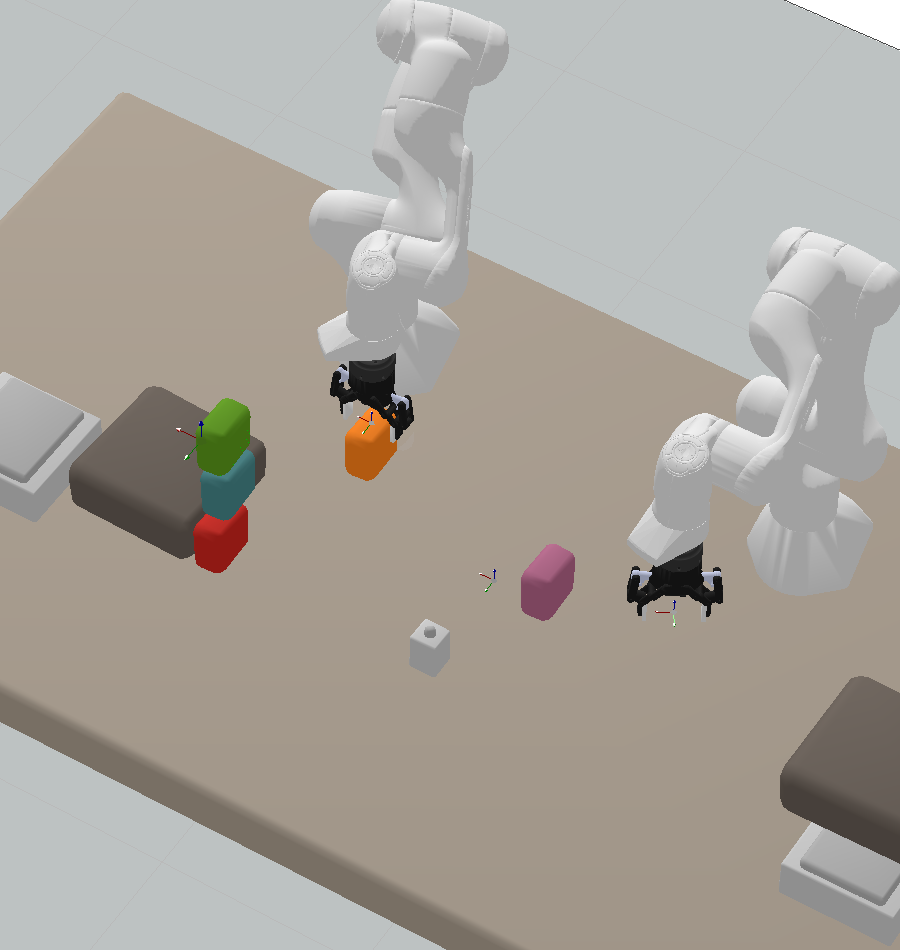} 
  \end{tabular}
  \caption{Manipulation sequence in \textit{+ Actions}. Robots build a tower [\textit{red}, \textit{blue}, \textit{green}, \textit{orange}], moving first an obstacle.}
  \label{fig:view_skeleton}
  \vspace{-0.6cm}
\end{figure}

\subsection{Integration in a Conflict-based TAMP Planner}

We demonstrate the benefits of neural accelerated conflict extraction inside the 
GraphNLP Planner \cite{ortiz2022conflict} for solving TAMP problems. The planner 
iteratively generates high-level plans, detects infeasible subgraphs, and encodes this information back into the logical description of the problem.



For this evaluation, we define 10 high-level goals for each setting corresponding to the \texttt{+Actions}, \texttt{+Robots}, and  \texttt{+Blocks} datasets, and report the total sum (including the 10 goals) of the number of solved NLPs and the computational time in the conflict extraction component of TAMP solver. \textit{GNN+e} (which is more robust than \textit{GNN+g1} in this setting) takes
only (8.33s, 511 NLPs), (9.83s, 603 NLPs), and (63.9s, 1979 NLPs) for each scenario, and is between 2 and 3 times faster than the \textit{expert} algorithm, which requires (24.2s, 731 NLPs), (38.7s, 1116 NLPs), and (137.9s, 2554 NLPs).





\section{Conclusion}


In this paper, we have presented a neural model to predict the minimal infeasible subsets of variables and constraints in a factored nonlinear program. The structure of the nonlinear program is used for neural message passing, providing generalization to problems with more variables and constraints.

We have demonstrated our approach in manipulation planninFg. A single learned model, combined with a suitable NLP representation of the motion sequence, can predict minimal infeasibility of manipulation sequences of different lengths in different scenes, increasing the number of objects and robots.
Our model achieves high accuracy, and the predictions can be integrated to guide and accelerate classical and heuristic algorithms for detecting minimal conflicts.

As future work, we would like to apply our neural formulation to detect conflicts in discrete constraint satisfaction problems, such as Boolean Satisfaction or $k$-coloring. From a robotics perspective, 
we will further investigate the potential of 
graph neural networks to combine logic and geometric information for guiding task and motion planning.










\newpage

\bibliographystyle{IEEEtran}
\bibliography{IEEEabrv,IEEEexample}

\end{document}